\documentclass[11pt,a4paper]{article}
\usepackage[nohyperref]{acl2020}  
\pdfoutput=1
\usepackage{times}
\usepackage{latexsym}

\usepackage{amssymb}
\usepackage{pifont}
\usepackage{multirow}
\usepackage{booktabs}
\usepackage{color}
\definecolor{red}{rgb}{1,0,0}
\definecolor{green}{rgb}{0,1,0}
\definecolor{bleu}{rgb}{0,0,1}

\newcommand{\xmark}{\ding{55}}
\usepackage{float}
\restylefloat{table}
\usepackage{paralist}

\usepackage{microtype}

\title{CLUECorpus2020: A Large-scale Chinese Corpus for Pre-training Language Model}

\date{}

\begin{document}

\maketitle
\begin{abstract}
In this paper, we introduce the Chinese corpus from CLUE organization, CLUECorpus2020, a large-scale corpus that can be used directly for self-supervised learning such as pre-training of a language model, or language generation. It has 100G raw corpus with 35 billion Chinese characters, which is retrieved from Common Crawl\footnote{\urlstyle{same}\url{http://commoncrawl.org}}. 
To better understand this corpus, we conduct language understanding experiments on both small and large scale, and results show that the models trained on this corpus can achieve excellent performance on Chinese.
We release a new Chinese vocabulary (vocab\_clue) with a size of 8K, which is only one-third of the vocabulary size used in Chinese Bert released by Google. It saves computational cost and memory while works as good as original vocabulary. We also release both large and tiny versions of the pre-trained model on this corpus. The former achieves the state-of-the-art result, and the latter retains most precision while accelerating training and prediction speed for eight times compared to Bert-base. To facilitate future work on self-supervised learning on Chinese, we release our dataset, new vocabulary, codes, and pre-trained models on Github\footnote{\urlstyle{same}\url{https://github.com/CLUEbenchmark/CLUECorpus2020/}}.


\end{abstract}
\section{Introduction}
\label{sec:introduction}

Transfer learning in natural language processing (NLP), which firstly pre-training a large model on the raw text, then fine-tuning on downstream tasks, now becomes the mainstream paradigm. It leverages large-scale raw text, which is abundant on the internet and achieves excellent performance. For example, T5~\citep{2019t5} treats all NLP problems as ``text-to-text" problem, trained on Colossal Clean Crawled Corpus (C4) with 750 GB raw corpus, and achieves the state-of-the-art performance on GLUE~\citep{wang2018glue} and SQuAD~\citep{rajpurkar2016squad}. 

Behind the rapid development of NLP recently, new and better models become available, and large scale raw corpus become more and more critical. There are several large-scale pre-training datasets publicly available in English. However, there is still a lack of open source large-scale Chinese corpus that can serve for the pre-training of language model. Therefore, we release CLUECorpus2020. For the convenience of reference, we also name it as C5, which stands for Colossal Clean Crawled Corpus for Chinese.

It contains 100 GB Chinese raw corpus, which is retrieved from Common Crawl. It is a well-defined dataset that can be used directly for pre-training without requiring additional pre-processing. CLUECorpus2020 contains around 29k separate files with each file following the pre-training format for the training set. And it has some amount of files for the development and test set, but each set is smaller. And several experiments on this dataset have been conducted to test the quality of this dataset.

To summarize, this paper makes the following contributions:
\begin{itemize}
\setlength\itemsep{-0.5em}
    \item A large-scale Chinese raw corpus that can be used for pre-training, language generation or learning word representation, such as word embedding.
    \item Through our experiments, we show that the model trained on a small percentage of our corpus can achieve better performance than the model trained on Chinese Wiki, which indicates the excellent quality and big potential of the dataset. With the whole dataset, we are able to match the state-of-the-art result on Chinese.
    \item A compact vocabulary( vocab\_clue) that can be used for NLP tasks in Chinese with only 8k vocabulary size, which is one-third of the vocabulary size of Chinese Bert( vocab\_bert). Models trained on vocab\_clue and vocab\_bert achieve comparable performance, while our vocabulary is smaller and better suit for Chinese, and can be faster for training machine learning models.
    \item We also release large and tiny versions of our pre-trained models trained on this dataset. The large version achieves the state-of-the-art performance, while the tiny version can be used to accelerate experiments and real applications.  
\end{itemize}
\section{Related work}
\label{sec:related work}

For English, there are a large number of open-source unlabeled corpora. For example, 
\begin{inparaenum}[\it 1)]
\item Toronto Books Corpus~\citep{zhu2015aligning}, a 4 GB dataset contains text extracted from eBooks, which represents a diﬀerent domain of natural language. 
\item WebText-like~\citep{radford2019language}, a 17 GB WebText-like English dataset, only uses content from web pages that were submitted to the content aggregation website Reddit and received a “score” of at least. 
\item English Wikipedia, a 16 GB English Wikipedia text data which consists of millions of encyclopedia articles written collaboratively and can be found in  TensorFlow Datasets\footnote{\urlstyle{same}\url{https://www.tensorflow.org/datasets/catalog/wikipedia/}}, which omits all markup and reference sections from the articles. 
\item C4~\citep{2019t5} dataset, a 750 GB English dataset consists of hundreds of gigabytes of clean English text scraped from the web. 
\end{inparaenum}

But for Chinese, similar corpus collections are still relatively rare and have a small size. For example, 
\begin{inparaenum}[\it 1)]
\item THUCTC~\citep{sum2016thuctc}, a 2.19 GB dataset contains 740,000 news documents. 
\item Chinese Wikipedia\footnote{\urlstyle{same}\url{https://github.com/brightmart/nlp_chinese_corpus/}}, a 1.1 GB dataset contains Chinese Wikipedia text data. 
\end{inparaenum}
As we all know, the size of the existing Chinese dataset is relatively small. In this paper, to solve the problem of lacking large-scale unlabeled corpus in Chinese, we leverage Common Crawl which is crawled from the whole internet and pre-process this dataset in detail. Finally, we provide a bigger and higher-quality all-inclusive corpus.

\section{Dataset Description}
\label{sec:dataset description}

\begin{table}[ht]
\centering
\small
\begin{tabular}{c|c|c|c} 
\toprule
 Dataset & Token~(B) &Sentences~(M) &Size~(GB) \\  
\midrule
Train  & 34.7 & 106 & 99.0 \\
\midrule
Dev  & 0.18 & 3.9 & 0.5  \\
\midrule
Test & 0.18 & 3.9 & 0.5\\
\bottomrule
\end{tabular}
\caption{Statistical information of CLUECorpus2020. ``B": billion; ``M": million.
Dev and Test set were drawn from the same distribution of training set, which can be used to check the generalization ability of a model. e.g., check mask Language Model(LM) accuracy of a  model during the training stage, or perplexity of a language model after training.  }
\label{table:stat}
\end{table}

Before we release this corpus, there is few large-scale high-quality Chinese dataset designed for pre-training language model in Chinese. This corpus is around 100 GB and comes from different websites \footnote{\urlstyle{same}\url{http://commoncrawl.org/the-data/get-started/}}. We use the ratio of $99:0.5:0.5$ to split the data into the training set, development set and test set randomly. As we seen from samples, it covers all sorts of topics, like news, entertainment, sports, health, international affairs, movies, celebrities, and so on. We follow pre-training format to organize the files of our dataset: one sentence per line and add an empty line at the end of a document. The overall statistics of this corpus is described in Table \ref{table:stat}. We will elaborate on the data construction process in the next section.


\section{Dataset Construction}
\label{sec:dataset contruction}

Unlabeled large scale datasets for unsupervised learning play an increasingly important role for Chinese NLP tasks. We believe that higher-quality data will have a better impact on Chinese NLP tasks. In this paper, we select and provide high-quality unlabeled dataset. To generate datasets that satisfy our requirements, we leverage Common Crawl as a source of text scraped from the web. 

Common Crawl is an organization that crawls the web and freely provides its archives and datasets to the public. Common Crawl usually crawls internet web content once a month. Common Crawl's web archives consist of petabytes of data collected since 2011. First, we extract text content from the scraped HTML files according to the detailed rules. Unfortunately, the majority of text content contains gibberish like dirty text or source code that we think useless to NLP tasks in Chinese. Furthermore, the scraped text includes a lot of duplicate content. To solve these problems, we do further filtering and extraction using the following heuristics rules, by which we have special treatment for Chinese, in addition to referring to the filtering method of C4:

\begin{itemize}
  \setlength\itemsep{-0.5em}
  \item Since we focus on Chinese tasks, we select sentences whose language type is Chinese, if a language is mentioned.
  \item To avoid incomplete sentences, we remove characters from the end of the text, until we find a Chinese terminal punctuation mark (i.e., a period, question mark, or the end of double quotation mark).
  \item Since Chinese words that contain ``List of Dirty, Naughty, Obscene or Other Bad Words.”\footnote{\urlstyle{same}\url{https://github.com/LDNOOBW/List-of-Dirty-Naughty-Obscene-and-Otherwise-Bad-Words/}} have a bad effect on building a healthy and civilized internet environment, so we remove all sentences that contain them.
  \item Warnings state that Javascript should be enabled unlikely to be helpful for NLP tasks, so we remove any line with the word Javascript or JavaScript.
  \item To deduplicate the dataset, we discard all but one of any four-sentence span occurring more than once in the dataset.
  \item We replace consecutive blank characters (i.e., tabs, spaces, invisible characters, etc.) that are generally meaningless in sentences with space.
  \item Since the curly bracket ``\{” appears in many programming languages (such as Javascript, widely used on the web) but not in the natural text, we remove any sentence that contains a curly bracket.
  \item To generate pre-trained format data, we use pyltp~\citep{che2010ltp} to separate text content into sentences, one complete sentence per line. 
  \item Since too short sentences that may be problematic or incomplete sentences are not suitable for language model training, we only retain sentences longer than 5.
\end{itemize}
 



We download the corpus from July to December 2019 from Common Crawl. After the aforementioned ﬁltering method, we extract the corpus of 100 GB. The corpus is much larger than the previous most datasets used for pre-training (about 100 GB) and is clean and natural Chinese text.

\section{Creation of CLUE Vocab }
\label{sec:Vocab}

\begin{table}[ht]
\centering
\small
\begin{tabular}{c|c|c} 
\toprule
 Token Type & Google &CLUE \\  
\midrule
Simplified Chinese & 11378 & 5689\\
\midrule
Traditional Chinese &3264 & \xmark\\
\midrule
English & 3529& 1320\\
\midrule
Japanese  & 573 & \xmark \\
\midrule
Korean  &84 & \xmark\\
\midrule
Emoji  & 56 & \xmark\\
\midrule
Numbers &1179 &140\\
\midrule
Special Tokens &106 &106\\
\midrule
Other Tokens & 959 & 766\\
\midrule
Total & 21128 & 8021 \\
\bottomrule
\end{tabular}
\caption{Statistical information of the two versions of the dictionaries. ``Special Tokens" include ``[PAD]", ``[UNK]", ``[CLS]", ``[SEP]", ``[MASK]", ``$\langle$S$\rangle$", ``$\langle$T$\rangle$" and 99 unused tokens.} 
\label{table:vocab}
\end{table}

The original BERT model uses character-based tokenization for Chinese. But there are many redundant tokens in the original vocabulary. Therefore, we have compiled a refined vocabulary through automated scripts and manual review. A detailed comparison of the two versions of the dictionaries can be seen in Table \ref{table:vocab}. We remove many unnecessary tokens, which will not be used in most cases of Chinese NLP tasks, such as traditional Chinese, Japanese, Korean and emojis. For English, we remove the most prefix tokens except for the single character and retain the most suffix tokens to guarantee the tokenization for English words. Similarly, for tokens standing for numbers, we only keep separate numeric symbols and the more commonly used words that represent the year. In addition, we also remove tokens composed of more than two special symbols.

As a result, the vocabulary size is only one-third of the original size of Chinese BERT. We call it ``vocab\_clue".

\section{Experiments}
\label{sec:experiments}

\begin{table*}[ht]
\centering
\small
\begin{tabular}{c|c|c|c|c|c|c|c|c|c} 
\toprule
Index & Model & Vocab & Data & Steps & AFQMC &TNEWS &IFLYTEK & CMNLI& AVG \\  
\midrule
1&BERT-base & Google& Wiki (1 GB) & 125K &69.93 & 54.77& 57.54& 75.64&64.47 \\
 \midrule
2&BERT-base & Google &C5 (1 GB) & 125K &69.63 &55.72 & 58.87& 75.75& 64.99\\
 \midrule
 3&BERT-base & CLUE &C5 (1 GB) & 125K &69.00 &55.04 & 59.07& 75.84&64.74 \\
\midrule
4& BERT-base\dag & Google &C5 (1 GB)& 125K &69.57 &55.17 & 59.69& 75.86& 65.07\\
\bottomrule
\end{tabular}
\caption{Performance of Bert Models on CLUE benchmark (\urlstyle{same}\url{http://www.cluebenchmark.com}) . BERT-base\dag~stand for BERT-base mm. For each experiment, we select the best model using dev set after training stage, then submit to CLUE benchmark and get score. Performance comparison for different corpus through Index 1 and 2. Performance comparison of two vocabularies through Index 2 and 3. Comparison of attention mechanism through Index 2 and 4. BERT-base mm, stand for BERT-base with minus and element-wise multiplication.  }
\label{table:bert_models_1}
\end{table*}

\begin{table*}[ht]
\centering
\small
\begin{tabular}{c|c|c|c|c|c|c|c|c|c} 
\toprule
Index & Model & Vocab & Data & Steps & AFQMC &TNEWS &IFLYTEK & CMNLI& AVG \\  
\midrule
5&BERT-base & Google &C5 (1 GB) & 375K & 69.85 & 55.97 & 59.62 & 76.41 & 65.46 \\
 \midrule
 6&BERT-base & CLUE &C5 (1 GB)& 375K & 69.93 & 56.38 & 59.35  & 76.58 & 65.56 \\
  \midrule
 7&BERT-base & Google &C5 (3 GB)& 375K & 70.22 & 56.41 & 59.58  & 76.70 & 65.73 \\
  \midrule
8&BERT-base & CLUE &C5 (3 GB)& 375K & 69.49 & 55.97 & 60.12  & 77.66 & 65.81 \\
\bottomrule
\end{tabular}
\caption{The effects of more training data of C5 and steps. With three times steps(3* 125k), BERT-base model trained on C5, gain 0.47 to 0.82 point compare to same model trained with 125k using different vocabularies. With tree times training data(3* 1GB), BERT-base model gain 0.74 to 1.07 point compare to same model trained with 1GB data and 125K, 0.25 to 0.27 point compare to same steps on 1GB. And the model trained with CLUE vocab always better than with Google vocab. }
\label{table:bert_models_1point5}
\end{table*}


 
\begin{table*}[!htbp]
\centering
\small
\begin{tabular}{c|c|c|c|c|c|c} 
\toprule
 Index & Model & Vocab & Training Data & Steps  &ACC of Masked LM &Loss of Masked LM  \\ 
\midrule
 1 & BERT-base & Google & Wiki(1 GB)  & 125K & 72.24\% &1.2321 \\
 \midrule
2 & BERT-base & Google & C5 (1 GB)  & 125K & 77.94\%& 0.9702 \\
 \midrule
 3 & BERT-base & CLUE &  C5 (1 GB)  & 125K & 76.47\%& 1.0691 \\
  \midrule
4 & BERT-base mm & Google &  C5 (1 GB) & 125K & 78.02\%&0.9816 \\
\bottomrule
\end{tabular}
\caption{Training metrics of Bert Models. Accuracy and loss of masked LM is on training set. }
\label{table:bert_base}
\end{table*}
 
 \begin{table*}[!htbp]
\centering
\small
\begin{tabular}{c|c|c|c|c|c} 
\toprule
 Task  & Length &Batch Size &Learning Rate&Epoch & Save Steps \\ 
\midrule
 AFQMC &128 &16 &2e-5&3&300\\
 \midrule
 TNEWS &128 & 16&2e-5&3&300\\
 \midrule
 IFLYTEK &128&32&2e-5&3&300\\
 \midrule
 CMNLI &128&64&3e-5&2&300\\
\bottomrule
\end{tabular}
\caption{Hyper-parameters of fine-tuning on CLUE tasks. We keep all hyper-parameters the same throughout all the experiments.}
\label{table:bert_para}
\end{table*}

\begin{table*}
\centering
\small
\begin{tabular}{c|c|c|c|c|c}
\toprule
 Model & Vocabulary  & Vocabulary Size & Parameters & Training Device & Training Speed\\
\midrule 
 Bert-base  & google\_vocab & 21128 &102M & TPU V3-8 & 1000steps/404s\\
\midrule 
 Bert-base  & clue\_vocab  & 8021~($\downarrow$~62.04\%) & 92M~($\downarrow$~9.80\%) & TPU V3-8 &1000steps/350s~($\uparrow$~15.43\%)\\
\midrule 
 RoBERTa-tiny-clue  & clue\_vocab  & 8021~($\downarrow$~62.04\%) & 7.5M~($\downarrow$~92.6\%) & TPU V3-8 &1000steps/50s~($\uparrow$~708.0\%)\\
\bottomrule
\end{tabular}
\caption{Detailed speed comparison of ``google\_vocab" and ``clue\_vocab", and RoBERTa-tiny-clue with Bert-base. CLUE vocabulary is one-third of Google vocabulary, with ten percentage or more speedup. RoBERTa-tiny-clue is 7 to 8 times faster than Bert-base}
\label{table:comparison}
\end{table*}


 \subsection{Pre-training with CLUECorpus2020 and Wiki}
 In this section, we want to compare our new dataset with Wiki using the same model. We choose BERT~\cite{devlin2018bert} as our baseline model. We pre-train the BERT-base model with Wiki data and C5 data, respectively. Due to the limited computing resources, we have designed a comparison on a small-scale corpus. The performance on large-scale corpus will be added in the next version. To make a fair comparison, we keep the parameters of both models the same. Meanwhile, the size of Wiki and the selected part of C5 data are both 1 GB. We release both of this corpus to make our results reproducible. As the length of most classification tasks is less than 128, we set the maximum sequence length of the pre-training to 128. It also improves the speed of pre-training. We pre-train BERT using masked language model (LM) prediction task without the next sentence prediction (NSP) task, as NSP task makes the performance of models worse observed in some recent papers, such as RoBERTa ~\citep{liu2019roberta}.

Classification board of CLUE benchmark
comprises 6 tasks meant to test general language understanding ability. We use the following four tasks to test the performance of our models as the sequence length of these tasks is 128 or less. During fine-tuning CLUE benchmark tasks, we also maintain the same parameters, as can be seen in Table \ref{table:bert_para}. Here are four tasks we used, including different kinds of tasks.  

\begin{itemize}
  \setlength\itemsep{0em}
  \item \textbf{Sentence Pair Similarity}: 
  
  AFQMC\footnote{\urlstyle{same}\url{https://dc.cloud.alipay.com/index\#/topic/intro?id=3}} 
  
  \item \textbf{Sentiment Analysis}: 
  
  TNEWS\footnote{\urlstyle{same}\url{https://github.com/fatecbf/toutiao-text-classfication-dataset/}}  
  
  \item \textbf{Multi-label Classification}: 
  
  IFLYTEK~\citep{2019iflytek}
  
  \item \textbf{Natural Language Inference}: 
  
  CMNLI~\citep{conneau2018xnli,williams2017broad}
\end{itemize}
 
 As we can see in Table \ref{table:bert_models_1}, the performance of the model pre-trained on C5 is 0.52\% higher than the model pre-trained on Wiki. It suggests that our data quality is similar or even better than Wiki. As we only use one percent of the whole dataset, there is a big potential to have a good performance using the whole dataset.

\begin{table*}[ht]
\centering
\scriptsize
\begin{tabular}{c|l|c|c|c|c|c|c|c} 
\toprule
Index & Model & Vocab & Data &  AFQMC &TNEWS &IFLYTEK & CMNLI& AVG \\  
\midrule
9 & BERT-base~\citep{devlin2018bert} & Google & / &  73.70 &56.58 &60.29 & 79.69& 67.57 \\ \midrule
10  &  ALBERT-tiny & Google & (30GB) &  69.92 & 53.35 & 48.71 & 70.61 & 60.65 \\
  \midrule
11 & ELECTRA-joint-generator-tiny~\citep{clark2019electra} & Google  & / &  69.90 &54.63 &52.31 & 73.17 & 62.50 \\ 
 \midrule
12 &  RoBERTa-tiny-clue ~\citep{chinese-bert-wwm} & CLUE & C5 (100 GB) & 69.52 & 54.57 & 57.31 & 73.1 & 63.60 \\ 
\bottomrule
\end{tabular}
\caption{Performance of the tiny version of pre-training models. Our tiny version model, RoBERTa-tiny-clue, retains most of precision compared to BERT-base, only 4 points lower than BERT-base, while performance is much better than ALBERT-tiny. All scores were reported by submitting to CLUE benchmark.}
\label{table:bert_models_2_tiny}
\end{table*}

 \subsection{Comparison of Attention Mechanisms with C5}

 The backbone of pre-trained models, typically like BERT and its variants, is Transformer model~\citep{vaswani2017attention}. The key component is the self-attention mechanism. With our new dataset, we are able to explore some variants of this mechanism. A self-attention module takes in $n$ inputs and returns $n$ outputs. The self-attention mechanism allows the inputs to interact with each other (``self”) and find out what they should pay more attention to (``attention”). The outputs are aggregates of these interactions and attention scores.\footnote{\urlstyle{same}\url{https://towardsdatascience.com/illustrated-self-attention-2d627e33b20a}}
 
 We believe there is some room for improvement of attention mechanism, especially that current heavily used self-attention may still too simple and naive to represent the importance of information for the input sequence. So we try a variant of self-attention mechanism, as follows:

\begin{itemize}
  \setlength\itemsep{-0.5em}
  \item \textbf{BERT-base mm} (Minus and element-wise Multiplication): Given two vector, we want to use a simple and inexpensive computation method to compute the similarity of these two vectors, such as using the absolute minus and element-wise multiplication operation. Then we transform the result with a dense layer and add them to the attention score.

\end{itemize}

As we can see from Table \ref{table:bert_models_1} and Table \ref{table:bert_base}, our variant has similar or slightly better performance as/than baseline.
These results indicate that possible improvement of attention mechanisms may improve performance on downstream tasks. With our new dataset, researchers of NLP can explore their ideas regarding this area.
 
 \subsection{Performance Comparison of Two Vocabularies}
 In order to verify the rationality of our vocabulary, we train the BERT-base model using the original vocabulary published by Google and our refined new vocabulary. We use the same model, and keep all hyper-parameters the same, and the only difference is the vocabulary. Similar to the previous section, we used multiple downstream tasks, four tasks, to confirm the performance of the vocabulary. As can be seen from Table \ref{table:bert_models_1}, for Index $2$ and $3$, the performance is similar, with only 0.25 point difference. We believe that our new vocabulary, ``vocab\_clue" can be used in downstream NLP tasks in Chinese in the future, especially for those situations with limited resource and computation power.
 
 In table \ref{table:comparison}, we make a detailed comparison. As we can see,  the size of clue\_vocab is 62.04\% less than origin vocabulary and has about 9.84\% fewer parameters compared to BERT-base. We find the training speed is 15.43\% faster than the original BERT-base, which pre-trained both on TPU V3-8. 
\subsection{More training data of C5 and steps}
As we can see from Tabel \ref{table:bert_models_1point5} , under the same steps, the performance of BERT-base which use 3 GB of data is 0.27 point higher than the performance of BERT-base which use 1 GB of data through Index 5 and 7. Meanwhile, the performance of BERT-base which use clue\_vocab is 0.1 point higher than the performance of BERT-base which use google vocabulary through Index 5 and 6. It can be concluded that the increase of the corpus can improve the performance of the model, and our clue\_vocab is better than google vocabulary while BERT-base model is trained enough steps.

\subsection{Performance of Large Version}
We generate our training data the same as RoBERTa~\citep{liu2019roberta}, and remove the Next Sentence Prediction(NSP) task. To compare with RoBERTa-wwm-large\footnote{\urlstyle{same}\url{https://github.com/ymcui/Chinese-BERT-wwm}}, which is currently the best Chinese model, we also use whole word mask as our mask strategy.

To speed up pre-training in our experiments, similar to BERT, we first train 500k steps on 128 sequence length with batch size 8k. We then train 600k on 512 sequence length with batch size 4k, and make the model more suitable for tasks with longer sequence lengths.

As we can see from Table \ref{table:bert_models_2}, As the number of training steps increases, the performance of the model is gradually getting better. Finally, we can see our performance is better than the original roberta-wwm-large.

\subsection{Performance of Tiny version}
State-of-the-art models, like BERT, can achieve very good performance compared to other models. However, as these models are very big and deep, with hundreds of millions of parameters, they are usually very slow during the prediction stage. To ease this problem, we release a small version of the pre-trained model. We want to keep it as small and fast as possible but retain the most precision. For those tasks that are not too difficult, like classification with few labels or sentence-pair, we recommend using this small version model to replace those big and slow BERT models.

We name it RoBERTa-tiny-clue, as it bases on the model of RoBERTa, and trained with corpus and vocabulary from CLUE. We first train it on sequence length 128 for 500k steps with batch size 8k using 100G corpus, then we train it for additional 200k steps with the same batch size using an additional 30G corpus. Together, it trains with 5.6 billion training instances. 

The configuration of hyper-parameters is keep same as ALBERT-tiny \footnote{\urlstyle{same}\url{https://github.com/brightmart/albert_zh}}, with hidden size 312 for 4 layers. It is around ten times fast for training and prediction compare to BERT-base. It gains an additional ten percentage speed accelerate even compare to ALBERT-tiny as the vocabulary used, vocab-clue, is only one-third of the vocabulary of BERT. 
Most importantly, it's performance is much better than ALBERT-tiny. Check table \ref{table:bert_models_2_tiny} for performance comparison with Bert-base and ALBERT-tiny.

\subsection{Transfer Learning Among Similar Tasks using Pre-trained Models}

\begin{table*}[ht]
\centering
\scriptsize
\begin{tabular}{c|l|c|c|c|c|c|c|c|c|c} 
\toprule
Index & Model & Vocab & Data & Steps & Init & AFQMC &TNEWS &IFLYTEK & CMNLI& AVG \\  
\midrule
13 & RoBERTa-wwm-large' & Google & / & / & \xmark & 74.44 &58.41 &62.77 & 82.2& 69.46 \\ 
 \midrule
14 & RoBERTa-wwm-large' & Google & / & / & CMNLI& \textbf{75.19} &58.41 &62.77 & 82.2& 69.64 \\  
 \midrule
15 & RoBERTa-large-clue & CLUE & C5 (100 GB) &100K & \xmark& 69.9 &56.95 &62.08 & 80.48& 67.35 \\  
 \midrule
16 & RoBERTa-large-clue & CLUE & C5 (100 GB) & 200K& \xmark& 69.98 &58.66 &62.50 & 81.33& 68.12 \\  
 \midrule
17 & RoBERTa-large-clue & CLUE & C5 (100 GB) & 500K & \xmark& 74.00 &\textbf{58.70} &62.31 & 82.04& 69.26 \\  
 \midrule
18 & RoBERTa-large-clue & CLUE & C5 (100 GB) & 500K& CMNLI & 74.41 &\textbf{58.70} &62.31 & 82.04& 69.37 \\  
 \midrule
19 & RoBERTa-large-clue & CLUE  & C5 (100 GB) &650K & \xmark& 70.01 &58.52 &62.54 & \textbf{82.68}& 68.44 \\  
 \midrule
20 & RoBERTa-large-clue & CLUE & C5 (100 GB) & 650K & CMNLI & 74.41 &58.52 &62.54 & \textbf{82.68}& 69.54 \\  
 \midrule
21 & RoBERTa-large-clue & CLUE & C5 (130 GB) & 800K & CMNLI& 74.41 &58.38 &\textbf{63.58} & 82.36& \textbf{69.68} \\ 
\bottomrule
\end{tabular}
\caption{Performance of RoBERTa-large using 100G corpus and vocabulary from CLUE. Our model RoBERTa-large-clue trained with 100g achieve same performance as RoBERTa-wwm-large' \citep{chinese-bert-wwm}, slightly better performance in two tasks, IFLYTEK and CMNLI. Init with CMNLI means when training task AFQMC, the model is initialized from a model that trained on CMNLI. All scores were reported by submitting to CLUE benchmark.}
\label{table:bert_models_2}
\end{table*}

Pre-trained models are powerful, but it is still difficult for them to learn tasks without enough training data. We observe that the robust performance model RoBERTa-large-clue can not learn well on task AFQMC, a sentence-pair task.  The CMNLI task, which is also a sentence pair-task,  has a lot of training data( around 390k). Therefore, We first train CMNLI using our pre-trained model, then we use this trained model to initialize AFQMC. As a result, it is around 0.8 to 4 points of performance boost compare to initializing from the pre-trained model. See field of init with CMNLI on table \ref{table:bert_models_2}. We believe this is a kind of transfer earning, which uses knowledge learned from one task and applies it to another similar task. We name the model trained with CMNIL as  RoBERTa-pair, and release it on our repository. We believe for many other sentence-pair tasks, with the help of this model, people can also achieve better performance than initializing from general pre-trained models.


\section{Conclusion}
\label{sec:conclusion}

In this paper, we introduce CLUECorpus2020, a large-scale corpus that can be used directly for the pre-training language model in Chinese. It is the first well-defined large-scale public available dataset that serves the purpose of the pre-training language model in Chinese. We conduct experiments on a small portion of this new dataset and Chinese Wiki. The results prove that our dataset has good quality and huge potential. In addition, we conduct experiments on the full dataset for a full-network pre-trained model. We also release a new vocabulary which size is small but work well for Chinese tasks. With our corpus and vocabulary, our model is able to match state-of-the-art performance in Chinese. We also observe that transfer learning is useful among similar tasks, and it can boost performance. We release our dataset, vocabulary, pre-trained models, and codes on Github.

In this work, we focus on pre-training, especially for language understanding. However, this dataset can also be used for language generation and other NLP tasks. We leave these for further study.

\section{Acknowledgements}
\label{sec:acknowledgements}

Our research is supported by Cloud TPUs from Google's TensorFlow Research Cloud (TFRC). We thank Zhe Zhao, Junyi Li, Shaomian Zheng, Zhenzhong Lan , and Peng Li for the sharing fee of the experiments.

\bibliographystyle{acl_natbib}
\bibliography{acl2020}

\end{document}